\documentclass[sigconf]{acmart}
\usepackage{enumitem}
\usepackage{amsmath}
\usepackage{pifont}  
\AtBeginDocument{%
  }

\copyrightyear{2025}
\acmYear{2025}
\setcopyright{cc}
\setcctype{by}
\acmConference[SIGSPATIAL '25]{The 33rd ACM International Conference on Advances in Geographic Information Systems}{November 3--6, 2025}{Minneapolis, MN, USA}

\acmBooktitle{The 33rd ACM International Conference on Advances in Geographic Information Systems (SIGSPATIAL '25), November 3--6, 2025, Minneapolis, MN, USA}
\acmDOI{10.1145/3748636.3762724}
\acmISBN{979-8-4007-2086-4/2025/11}

\begin{document}

\title[Baltimore Atlas: Ultra-high Spatial Resolution Land Cover Classification]{Baltimore Atlas: FreqWeaver Adapter for Semi-supervised Ultra-high Spatial Resolution Land Cover Classification}

\author{Junhao Wu}
\email{jwu17@students.towson.edu}
\orcid{1234-5678-9012}
\affiliation{%
  \institution{Towson University}
  \city{Towson}
  \state{MD}
  \country{USA}
}

\author{Aboagye-Ntow Stephen}
\email{saboagy1@students.towson.edu}
\affiliation{%
  \institution{Towson University}
  \city{Towson}
  \state{MD}
  \country{USA}
}

\author{Chuyuan Wang}
\email{cwang@towson.edu}
\affiliation{%
  \institution{Towson University}
  \city{Towson}
  \state{MD}
  \country{USA}
}

\author{Gang Chen}
\email{Gang.Chen@charlotte.edu}
\affiliation{%
  \institution{University of North Carolina at Charlotte}
  \city{Charlotte}
  \state{NC}
  \country{USA}
}

\author{Xin Huang}
\email{xhuang@towson.edu}
\affiliation{%
  \institution{Towson University}
  \city{Towson}
  \state{MD}
  \country{USA}
}

\renewcommand{\shortauthors}{Junhao Wu et al.}

\begin{abstract}
Ultra-high Spatial Resolution (UHSR) Land Cover Classification is increasingly important for urban analysis, enabling fine-scale planning, ecological monitoring, and infrastructure management. It identifies land cover types on sub-meter remote sensing imagery, capturing details such as building outlines, road networks, and distinct boundaries. However, most existing methods focus on 1\,m imagery and rely heavily on large-scale annotations, while UHSR data remain scarce and difficult to annotate, limiting practical applicability. To address these challenges, we introduce \textit{Baltimore Atlas}, a UHSR land cover classification framework that reduces reliance on large-scale training data and delivers high-accuracy results. \textit{Baltimore Atlas} builds on three key ideas: (1) \textit{Baltimore Atlas Dataset}, a 0.3\,m resolution dataset based on aerial imagery of Baltimore City; (2) \textit{FreqWeaver Adapter}, a parameter-efficient adapter that transfers SAM2 to this domain, leveraging foundation model knowledge to reduce training data needs while enabling fine-grained detail and structural modeling; (3) \textit{Uncertainty-Aware Teacher Student Framework}, a semi-supervised framework that exploits unlabeled data to further reduce training dependence and improve generalization across diverse scenes. Using only 5.96\% of total model parameters, our approach achieves a 1.78\% IoU improvement over existing parameter-efficient tuning strategies and a 3.44\% IoU gain compared to state-of-the-art high-resolution remote sensing segmentation methods on the Baltimore Atlas Dataset.
\end{abstract}



\maketitle

\section{Introduction}

Ultra-high Spatial Resolution (UHSR) remote sensing imagery, typically at sub-meter scales, offers rich spatial detail that allows accurate identification of individual ground objects and preservation of their structural features~\cite{Lv_Shen_Lv_Li_Shi_Zhang_2023}. Compared with coarser imagery, it provides sharper boundaries, finer textures, and the ability to capture small, heterogeneous features that are essential for detailed geospatial analysis~\cite{Guo_2024_SkySense}. Leveraging these strengths, UHSR land cover classification converts high-detail imagery into thematic maps that combine broad spatial patterns with precise object-level information~\cite{gui2024remote}. The fine granularity of UHSR data reduces mixed-pixel effects, improves boundary accuracy, and enables reliable detection of subtle features such as narrow roads, isolated vegetation patches, and variations in roof materials or tree canopy structures. These advantages produce classification results that more closely match real-world surface conditions, supporting parcel-level land use mapping, high-precision urban modeling, ecosystem monitoring, and disaster impact assessment~\cite{li2024hierarchical}. By linking large-scale landscape observation with fine-scale object analysis, UHSR-based land cover classification has become an essential tool for modern geospatial applications.

Despite its importance, most current remote sensing segmentation methods still focus on imagery with 1-meter spatial resolution, which is sufficient for identifying coarse land cover categories such as large vegetation patches, water bodies, and major transportation corridors, but struggles to capture fine-grained details such as narrow alleyways, small rooftop structures, or individual vehicles~\cite{mendieta2024toward,ma2025transform,ZHANG2022113106}. UHSR imagery, by contrast, enables the recognition of such detailed structures and subtle boundaries, offering the potential for parcel-level urban analysis and more precise land cover mapping~\cite{huang2025multiscale}. However, annotated UHSR datasets remain scarce and are extremely time- and labor-intensive to produce~\cite{Liu_2025_TGRS,he2024annotated}, as pixel-level labeling at this resolution demands careful delineation of numerous small and heterogeneous objects. This annotation bottleneck has limited large-scale exploration of UHSR land cover classification in both research and practical applications.

To address these limitations, we propose \textit{Baltimore Atlas}, 
a novel framework for UHSR land cover classification.  
Baltimore Atlas performs multi-scale frequency-decoupled modeling for complex UHSR data patterns, preserving global boundaries while capturing fine-grained local details.
Moreover, the framework leverages the prior knowledge of foundation models 
and integrates a semi-supervised learning strategy to reduce reliance on large annotated datasets.  
Consequently, the framework achieves high-accuracy and robust UHSR land cover mapping under limited training data availability.

At the heart of \textit{Baltimore Atlas} is the \textit{FreqWeaver Adapter}, a frequency-modulated module designed to adapt SAM2 for UHSR imagery. To address the diverse spatial patterns in UHSR data, the module first applies spectral filtering to separate input features into low- and high-frequency components. The low-frequency branch employs large-receptive-field convolutions to capture global semantics, while the high-frequency branch applies small-receptive-field convolutions to preserve fine-grained structures. In parallel, a complementary spatial-domain residual branch restores local details. This design introduces only two additional convolutional kernels, maintaining a parameter-efficient architecture, and enables the model to effectively integrate global context with fine-grained detail for more accurate UHSR land cover classification.

Further, we introduce the \textit{Uncertainty-Aware Teacher Student Framework}, a semi-supervised framework for UHSR data. The teacher model takes weakly perturbed inputs and generates predictions. We compute pixel-wise entropy from these predictions, and lower entropy means higher confidence. For each pixel, the class with the highest confidence is chosen as the pseudo-label. High-confidence regions give stronger supervision to the student model, which is trained on strongly perturbed inputs. Focusing on these reliable regions helps the framework reduce noise from low-confidence areas, learn more stable representations, and improve performance and generalization in complex scenes. In addition, leveraging large-scale unlabeled data further reduces the reliance on annotated training samples.

\noindent The main contributions of this paper are summarized as follows:
\begin{itemize}[leftmargin=*, noitemsep, topsep=0pt]
    \item We propose \textit{Baltimore Atlas}, a novel framework for UHSR land cover classification that achieves high-accuracy and robust mapping even with limited training data.
    \item We design a unified architecture with two core components:  \textit{FreqWeaver Adapter}, which parameter-efficiently adapts SAM2 to complex UHSR data patterns, and \textit{Uncertainty-Aware Teacher Student Framework}, which uses entropy-based confidence to learn stable representations and reduce complex noise.
    \item We introduce the \textit{Baltimore Atlas Dataset}, a 0.3\,m resolution dataset based on aerial imagery of Baltimore City. Experiments on this dataset demonstrate consistent improvements over prior methods.
\end{itemize}

\section{Relate Work}

\subsection{Ultra-high Spatial Resolution Land Cover Classification}

Early land cover classification mainly used pixel-based methods, treating each pixel as an independent unit classified only by its spectral features. In high spatial resolution imagery, these methods often cause confusion and fragmentation due to the high spectral similarity among different land covers~\cite{roy2018comparative,han20}, leading to the development of Geographic Object-Based Image Analysis (GEOBIA/OBIA)~\cite{Chen04032018, Hay2008,rs12122012}. GEOBIA aggregates similar pixels into image objects and combines segmentation, feature extraction, and semantic rules to improve classification, but large-scale mapping still requires considerable expert knowledge and manual intervention. To reduce manual effort and improve automation, deep learning has been widely applied to fine-grained land cover classification. LSHR-LCM~\cite{Robinson_2019_CVPR} evaluated multiple deep learning models on 1~m resolution data across the United States, demonstrating the feasibility of convolutional networks for nationwide mapping but showing strong dependence on high-quality labeled data. ABCNet~\cite{Li_2021} improved network design to preserve fine spatial details while capturing global context, enhancing both accuracy and stability in high-resolution semantic segmentation. Urban Watch~\cite{ZHANG2022113106} refined the impervious surface class, resulting in a total of nine urban categories, to address material diversity and overly coarse definitions, and incorporated semi-supervised learning to reduce annotation requirements. LEG~\cite{10655508} used 30~m low-resolution labels for weak supervision, improving 1~m mapping accuracy and demonstrating the potential for cross-resolution transfer. TLULC~\cite{Khan2024} applied large-scale Transformer pretraining and transfer learning, further enhancing the generalization and robustness of high-resolution mapping. Despite these advances, most methods focus on 1~m or coarser resolutions, leaving UHSR land cover classification largely unexplored.

To address this gap, we propose the \textit{Baltimore Atlas}, a framework for UHSR(0.3~m resolution) land cover classification that combines tailored transfer learning with semi-supervised learning to adapt to the complex spatial patterns of UHSR imagery, achieving improved performance while reducing reliance on large-scale annotated datasets.

\subsection{Parameter-Efficient Adaptation for Multi Scale Remote Sensing Understanding}

Foundation models (FMs) such as SAM2~\cite{ravi2024sam2segmentimages} are trained on large-scale natural image datasets with segmentation labels, enabling strong generalization across segmentation tasks. DED-SAM~\cite{ded-sam2024} demonstrates SAM2’s zero-shot segmentation capability on remote sensing (RS) imagery. To better align model representations with RS characteristics, GeoFMs such as Clay~\cite{clay2024}, Prithvi~\cite{Prithvi}, and S2MAE~\cite{cvpr2024_s2mae} incorporate RS-specific priors through pretraining on satellite and aerial imagery. However, these models still lack task-specific priors for segmentation. As a result, both general-purpose FMs and GeoFMs require adaptation to downstream datasets to achieve strong performance in tasks like land cover classification. Parameter-efficient fine-tuning (PEFT) addresses this need by reducing trainable parameters while maintaining accuracy: LoRA~\cite{hu2021loralowrankadaptationlarge} constrains weight updates to low-rank subspaces, prompt tuning~\cite{jia2022visualprompttuning} and prefix tuning~\cite{li2021prefixtuningoptimizingcontinuousprompts} insert learnable vectors into the input or attention layers, and adapter tuning~\cite{NEURIPS2022_69e2f49a} adds lightweight modules inside transformer layers. While effective in many domains, these approaches often struggle with the multi-scale structures, rich spectral content, and complex semantics of RS data. To bridge this gap, RS-specific PEFT designs have been proposed: AiRs~\cite{AiRs2024} models spatial context and semantic responses, TEA~\cite{TEA2024} employs a lightweight side adapter with macro-scale semantic guidance, and Earth-Adapter~\cite{hu2025earthadapterbridgegeospatialdomain} applies frequency decoupling for multi-scale modeling but relies on a unified linear encoder, limiting scale-specific feature modeling.

To address challenges in remote sensing imagery, we propose the \textit{FreqWeaver Adapter}, a novel adapter design that integrates frequency-domain decomposition with convolutional processing. It separates high- and low-frequency components through dedicated convolutional branches and incorporates spatial information. This design enables efficient multi-scale adaptation, reduces aliasing artifacts, avoids large variations in object scale, while maintaining a very low additional parameter cost.

\subsection{Semi-Supervised Learning for Robust Remote Sensing Representations}

Semi-supervised learning (SSL) leverages both labeled and unlabeled data to improve model generalization and robustness under limited annotations. General SSL strategies include pseudo-labeling~\cite{lee2013pseudo, xie2020noisystudent}, consistency regularization~\cite{berthelot2019mixmatch, sohn2020fixmatch}, and teacher student frameworks~\cite{tarvainen2018meanteachersbetterrole}. While these approaches have demonstrated strong performance in natural image tasks, their direct application to remote sensing (RS) is often limited by challenges such as multi-scale spatial structures and high inter-class spectral similarity. To overcome these limitations, several methods have extended SSL to better fit RS characteristics: SSSM~\cite{feng2022sssm} introduced multi-level feature fusion with pseudo-label sampling for scene classification, alleviating the noise caused by low-confidence labels; UniMatch~\cite{yang2023unimatch} enhanced semantic segmentation with task-driven perturbation branches to better handle complex RS structures; MUCA~\cite{wang2025muca} incorporated multi-scale uncertainty consistency and cross teacher--student attention to align intermediate features across scales, improving the discrimination of spectrally similar land-cover classes; and SOOD~\cite{hua2023sood} adapted semi-supervised learning to oriented object detection by introducing orientation-consistency and global-consistency losses, significantly improving detection performance on aerial benchmarks. Overall, these methods highlight the potential of SSL to reduce annotation costs and enhance performance in complex RS scenarios, but their applications to UHSR data remain insufficiently explored.

To reduce the noise caused by unstable predictions in UHSR imagery, we propose the \textit{Uncertainty-Aware Teacher–Student Framework}. This framework uses entropy-based uncertainty estimation to select high-confidence pseudo-labels and focuses supervision on reliable regions. In this way, it limits the effect of noisy predictions, makes better use of large amounts of unlabeled data, and learns more stable feature representations while lowering dependence on extensive annotations.

\begin{figure*}[t]
    \centering
    \includegraphics[width=1\textwidth]{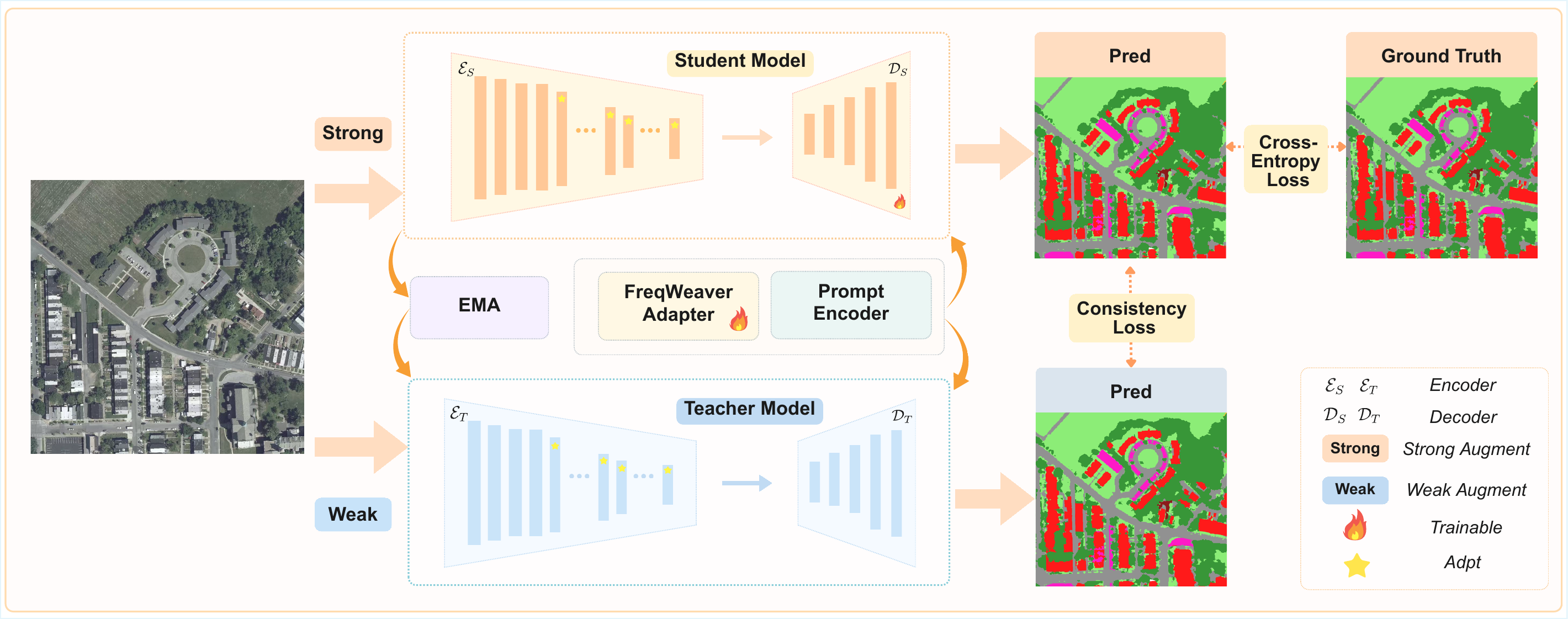} 
    \caption{Overview of the proposed framework. Given a remote sensing image, two augmented views are generated: a strongly augmented version is fed into the student network, and a weakly augmented version is processed by the teacher network. To capture the characteristics of UHSR data, a parameter-efficient FreqWeaver Adapter is inserted into specific Transformer layers. The teacher network is updated with an exponential moving average (EMA) strategy to maintain stable parameter evolution. For unlabeled data, the teacher network provides pseudo-labels to enforce consistency with student predictions, enabling more robust feature learning. For labeled data, ground-truth annotations directly supervise the student network to ensure accurate semantic understanding. Overall, the framework integrates adapter-based adaptation with semi-supervised learning to deliver reliable segmentation of UHSR imagery while reducing reliance on labeled data.}
    \label{fig:framework}
\end{figure*}

\section{Method}
\subsection{Overview}
We propose \textit{Baltimore Atlas}, a framework for ultra-high spatial resolution (UHSR) land cover classification. To support this framework, we construct the \textit{Baltimore Atlas Dataset}, a benchmark created from 0.3 m aerial imagery of Baltimore City acquired through the National Agriculture Imagery Program (NAIP). For the model, we adopt \textit{SAM2} as the foundation and introduce the \textit{FreqWeaver Adapter}, which enables parameter-efficient fine-tuning and adapts \textit{SAM2} to the multi-scale patterns of UHSR data. In addition, we design the \textit{Uncertainty-Aware Teacher–Student Framework}, a semi-supervised learning method that employs uncertainty estimation to generate high-confidence pseudo-labels, which are then used as supervisory signals for training on unlabeled data. Overall, the integration of these components enables accurate land cover classification on UHSR data with limited annotated samples, as illustrated in Fig.~\ref{fig:framework}.

\subsection{FreqWeaver Adapter}

\begin{figure*}[t]
    \centering
    \includegraphics[width=0.85\textwidth]{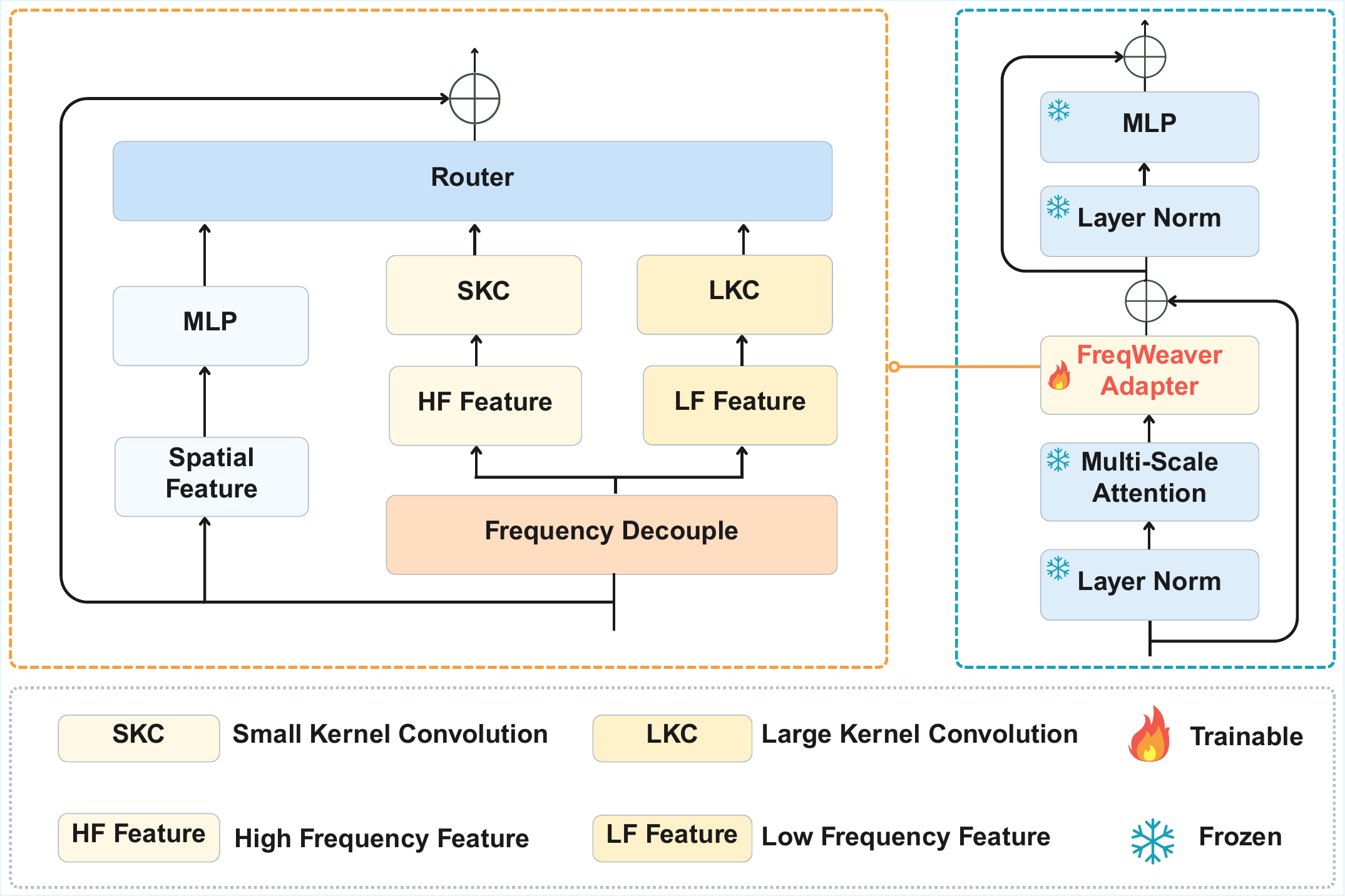}
    \caption{Overview of the proposed FreqWeaver Adapter. The FreqWeaver Adapter is inserted into specific Transformer layers as the only trainable component, while all other parameters remain frozen. It performs frequency decoupling by applying a Fourier transform to separate high- and low-frequency components. High-frequency information is processed with small convolution kernels to capture local textures and fine details, while low-frequency information is modeled with large kernels to enhance global boundary features. To prevent information loss, the original spatial features are preserved, and a router is employed to fuse multi-scale information effectively. This design provides more comprehensive feature representations and enables the model to better adapt to the multi-scale characteristics of UHSR data.}
    \label{fig:adapter}
\end{figure*}

A key challenge of UHSR imagery lies in its significant scale variation, dense object distribution, and complex textures, which limit the adaptability of \textit{SAM2} and existing parameter-efficient tuning (PEFT) methods. To address this, we introduce \textit{FreqWeaver}, a lightweight frequency–spatial modulation adapter that improves multi-scale modeling while keeping parameter cost low. The method integrates frequency-domain decomposition with convolutional processing: it separates low- and high-frequency components through dedicated convolutional branches and incorporates spatial information for more complete feature representation. Our approach is organized into three stages—\textit{Frequency Decomposition}, \textit{Multi-Branch Feature Modeling}, and \textit{Adaptive Feature Fusion}—which will be described in the following sections. An overview of the proposed adapter is illustrated in Figure~\ref{fig:adapter}.

\subsubsection{Frequency Decomposition}

To enable frequency-aware feature refinement, we apply frequency decomposition to the encoder features using the 2D Discrete Fourier Transform (DFT). This operation explicitly separates low- and high-frequency components for specialized processing.

Before applying the DFT, the input feature tensor is reshaped from $\mathbf{X} \in \mathbb{R}^{B \times H \times W \times C}$ to $\mathbf{X}' \in \mathbb{R}^{B \times C \times H \times W}$ to align with the channel-wise Fourier transform convention.

We then apply the 2D DFT independently to each channel as:
\begin{equation}
\mathbf{X}_{\mathrm{f}} = \mathcal{F}_{\mathrm{DFT}}(\mathbf{X}'),
\end{equation}
where $\mathcal{F}_{\mathrm{DFT}}$ denotes the 2D Fourier transform along the spatial dimensions $(H, W)$ for each channel. The resulting spectrum $\mathbf{X}_{\mathrm{f}} \in \mathbb{C}^{B \times C \times H \times W}$ contains complex-valued frequency coefficients.

To extract different frequency bands, two complementary radial binary masks are defined:
\begin{equation}
\mathcal{M}_{\mathrm{lf}} = \mathbb{I}(r < \rho), \quad 
\mathcal{M}_{\mathrm{hf}} = 1 - \mathcal{M}_{\mathrm{lf}},
\end{equation}
where $\mathbb{I}(\cdot)$ denotes the indicator function, $r$ is the radial distance from the spectral center, and $\rho$ is a predefined cutoff threshold.

The low- and high-frequency components are isolated by applying the masks element-wise:
\begin{equation}
\mathbf{X}_{\mathrm{lf}} = \mathbf{X}_{\mathrm{f}} \odot \mathcal{M}_{\mathrm{lf}}, \quad
\mathbf{X}_{\mathrm{hf}} = \mathbf{X}_{\mathrm{f}} \odot \mathcal{M}_{\mathrm{hf}},
\end{equation}
where $\odot$ denotes Hadamard (element-wise) multiplication.

Finally, the spatial-domain representations are recovered via the inverse Fourier transform:
\begin{equation}
\widehat{\mathbf{X}}_{\mathrm{lf}} = \mathcal{F}^{-1}_{\mathrm{DFT}}(\mathbf{X}_{\mathrm{lf}}), \quad
\widehat{\mathbf{X}}_{\mathrm{hf}} = \mathcal{F}^{-1}_{\mathrm{DFT}}(\mathbf{X}_{\mathrm{hf}}),
\end{equation}
yielding real-valued tensors $\widehat{\mathbf{X}}_{\mathrm{lf}}, \widehat{\mathbf{X}}_{\mathrm{hf}} \in \mathbb{R}^{B \times C \times H \times W}$ that represent the low- and high-frequency spatial features. These components are then forwarded to separate branches for frequency-specific refinement.

\subsubsection{Multi-Branch Feature Modeling}

To enhance frequency- and spatial-domain representations in a disentangled manner, FreqWeaver introduces three specialized branches, each responsible for modeling a specific component of the feature representations: low-frequency (global structure), high-frequency (local detail), and spatial-domain semantics. These branches operate independently and produce complementary signals for subsequent fusion.

\paragraph{Low-Frequency Branch.}
This branch targets the restoration of global semantic coherence and suppression of large-scale distortions, which are commonly observed in low-frequency-dominated regions such as terrain or vegetation. By modeling long-range dependencies, it reinforces region-level consistency across the scene.

The low-frequency response is computed as:
\begin{equation}
\mathbf{F}_{\mathrm{lf}} = f_{\mathrm{lf}}(\widehat{\mathbf{X}}_{\mathrm{lf}}),
\end{equation}
where $\widehat{\mathbf{X}}_{\mathrm{lf}} \in \mathbb{R}^{B \times C \times H \times W}$ is the input low-frequency feature map, and $f_{\mathrm{lf}}(\cdot)$ denotes a depthwise convolution with a kernel size of $11 \times 11$. The output $\mathbf{F}_{\mathrm{lf}} \in \mathbb{R}^{B \times C \times H \times W}$ encodes coarse structural information and is reshaped into $\mathbb{R}^{B \times H \times W \times C}$.

\paragraph{High-Frequency Branch.}
This branch emphasizes fine-grained details such as edges, contours, and textures—high-frequency components that are essential for distinguishing man-made objects and detailed structures. It enhances boundary clarity and preserves local precision.

The high-frequency response is computed as:
\begin{equation}
\mathbf{F}_{\mathrm{hf}} = f_{\mathrm{hf}}(\widehat{\mathbf{X}}_{\mathrm{hf}}),
\end{equation}
where $\widehat{\mathbf{X}}_{\mathrm{hf}} \in \mathbb{R}^{B \times C \times H \times W}$ is the high-frequency input, and $f_{\mathrm{hf}}(\cdot)$ is a depthwise convolution with a $5 \times 5$ kernel. The output $\mathbf{F}_{\mathrm{hf}} \in \mathbb{R}^{B \times C \times H \times W}$ highlights spatially localized structures and is reshaped into $\mathbb{R}^{B \times H \times W \times C}$.

\paragraph{Spatial Branch.}
This branch directly processes the spatial-domain features without frequency filtering. It is designed to recover residual semantics that may be overlooked by the low- and high-frequency branches, such as transitional regions, occluded objects, or mixed textures. The spatial-domain response is defined as:
\begin{equation}
\mathbf{F}_{\mathrm{s}} = f_{\mathrm{s}}(\mathbf{X}),
\end{equation}
where $\mathbf{X} \in \mathbb{R}^{B \times H \times W \times C}$ is the original feature map before frequency decomposition, and $f_{\mathrm{s}}(\cdot)$ is a two-layer bottleneck MLP. The output $\mathbf{F}_{\mathrm{s}} \in \mathbb{R}^{B \times H \times W \times C}$ provides complementary spatial semantics to the frequency-based branches.

\subsubsection{Adaptive Feature Fusion}

To integrate the outputs of the low-frequency, high-frequency, and spatial branches, we adopt a content-aware fusion strategy guided by global contextual information. This mechanism enables the model to dynamically balance the relative contributions of the three feature types on a per-sample basis.

We first compute a global context descriptor by performing global average pooling over the spatial dimensions:
\begin{equation}
\mathbf{g} = \frac{1}{HW} \sum_{h=1}^{H} \sum_{w=1}^{W} \mathbf{X}_{:,h,w,:} \in \mathbb{R}^{B \times C},
\end{equation}
where $\mathbf{X} \in \mathbb{R}^{B \times H \times W \times C}$ denotes the original spatial-domain feature map.

The descriptor $\mathbf{g}$ is then passed to a lightweight router MLP, which outputs normalized fusion weights:
\begin{equation}
[r_{\mathrm{s}}, r_{\mathrm{lf}}, r_{\mathrm{hf}}] = \operatorname{softmax}(f_{\mathrm{router}}(\mathbf{g})) \in \mathbb{R}^{B \times 3},
\end{equation}
where $f_{\mathrm{router}}(\cdot)$ denotes a two-layer MLP with softmax normalization to ensure $\sum_k r_k = 1$.

Finally, the three branch outputs are fused with the original features via weighted addition:
\begin{equation}
\mathbf{Y} = \mathbf{X} + r_{\mathrm{s}} \odot \mathbf{f}_{\mathrm{s}} + r_{\mathrm{lf}} \odot \mathbf{f}_{\mathrm{lf}} + r_{\mathrm{hf}} \odot \mathbf{f}_{\mathrm{hf}},
\end{equation}
where $\odot$ denotes broadcasted element-wise multiplication. The final output $\mathbf{Y} \in \mathbb{R}^{B \times H \times W \times C}$ preserves the original spatial resolution while incorporating frequency- and content-aware enhancements.

This fusion strategy adaptively balances complementary information sources, thereby improving the model’s ability to generalize to diverse spatial patterns in remote sensing imagery.

\subsection{Uncertainty-Aware Teacher–Student Framework}

To address the noise in pseudo-labels for semi-supervised remote sensing segmentation, especially in UHSR imagery, we propose the \textbf{Uncertainty-Aware Mean Teacher (UAMT)} framework. The method employs entropy-based uncertainty estimation to select high-confidence pseudo-labels as supervisory signals for unlabeled data, reducing the impact of uncertain predictions. By combining this mechanism with a perturbation-consistent teacher–student scheme, UAMT stabilizes pseudo-supervision and enables efficient adaptation of SAM2 under limited annotation.

\subsubsection{Mean Teacher Framework with Perturbation Consistency}

Our framework adopts a teacher–student architecture where the student network $\boldsymbol{\theta}_{\mathrm{S}}$ is optimized via gradient descent, while the teacher network $\boldsymbol{\theta}_{\mathrm{T}}$ acts as a temporal ensemble, updated using Exponential Moving Average (EMA) to ensure stable supervision:
\begin{equation}
\boldsymbol{\theta}_{\mathrm{T}}^{(t)} = \alpha\, \boldsymbol{\theta}_{\mathrm{T}}^{(t-1)} + (1 - \alpha)\, \boldsymbol{\theta}_{\mathrm{S}}^{(t)}, \quad \alpha = 0.99,
\end{equation}
where $\alpha$ is the EMA decay coefficient controlling the momentum of the update.

To enforce robustness under data augmentation, each unlabeled input $x_{u} \in \mathcal{D}_{u}$ is subjected to two augmentation pipelines: a weakly augmented version $\widetilde{x}^{\text{weak}}_{u}$ is fed to the teacher model, while a strongly augmented counterpart $\widetilde{x}^{\text{strong}}_{u}$ is provided to the student. The underlying assumption is that semantic predictions should remain consistent across these perturbations, even if the appearance varies. Accordingly, consistency regularization is applied to align the outputs of teacher and student networks under this augmentation mismatch.

\subsubsection{Entropy-Based Uncertainty Estimation}

Pseudo-labels in remote sensing segmentation are often noisy due to complex backgrounds, fine boundaries, and domain-specific variations. To reduce their negative impact, we apply pixel-wise confidence weighting based on Shannon entropy.

Let $\mathbf{P}^{\text{T}}_{j} \in \mathbb{R}^{C}$ be the teacher's softmax probability vector at pixel $j$, where $C$ is the number of classes. The uncertainty is measured by entropy:
\begin{equation}
\mathcal{H}_{j} = - \sum_{c=1}^{C} \mathbf{P}^{\text{T}}_{j,c} \log \mathbf{P}^{\text{T}}_{j,c},
\end{equation}
which reaches $\log C$ when the distribution is uniform. We normalize this score and convert it to a confidence weight:
\begin{equation}
w_{j} = 1 - \frac{\mathcal{H}_{j}}{\log C}, \quad w_{j} \in [0,1],
\end{equation}
where higher $w_{j}$ values indicate more reliable predictions and thus stronger supervision. This scheme downweights uncertain regions and emphasizes high-confidence pseudo-labels.

\subsubsection{Loss Function: Supervised–Unsupervised Coupling}

The training objective combines a supervised loss on labeled data and a confidence-weighted consistency loss on unlabeled data.

\paragraph{Supervised Loss.}
For labeled pairs $(x_{l}, y_{l}) \in \mathcal{D}_{l}$, the student is trained with pixel-wise cross-entropy:
\begin{equation}
\mathcal{L}_{\mathrm{sup}} = \mathrm{CE}\!\left(f(x_{l}; \boldsymbol{\theta}_{\mathrm{S}}), y_{l}\right),
\end{equation}
where $f(\cdot;\boldsymbol{\theta}_{\mathrm{S}})$ denotes the student prediction.

\paragraph{Consistency Loss.}
For unlabeled inputs $x_{u} \in \mathcal{D}_{u}$, the teacher produces probabilities $\mathbf{P}^{\text{T}}$, from which a pseudo-label $\hat{y}^{\text{T}}_{j} = \arg\max_{c} \mathbf{P}^{\text{T}}_{j,c}$ is obtained per pixel. The student predicts $\mathbf{P}^{\text{S}}$ from a perturbed view. Using confidence weights $w_{j}$ derived from entropy, the consistency loss is:
\begin{equation}
\mathcal{L}_{\mathrm{cons}} = \frac{1}{B \cdot N} \sum_{x_{u} \in \mathcal{D}_{u}} \sum_{j=1}^{N} 
w_{j} \cdot \mathrm{CE}(\mathbf{P}^{\text{S}}_{j}, \hat{y}^{\text{T}}_{j}),
\end{equation}
where $B$ is the batch size and $N$ is the number of pixels per image.

\paragraph{Total Objective.}
The final loss is:
\begin{equation}
\mathcal{L}_{\mathrm{total}} = \mathcal{L}_{\mathrm{sup}} + \lambda \mathcal{L}_{\mathrm{cons}},
\end{equation}
with $\lambda = 0.1$ controlling the balance between labeled and unlabeled data.

During training, only the parameters of FreqWeaver adapters are updated, while the SAM2 backbone remains frozen, ensuring efficient transfer to remote sensing domains.

\section{Experiment}

\subsection{Dataset}
\begin{figure}[htbp]
    \centering
    \includegraphics[width=0.48\textwidth]{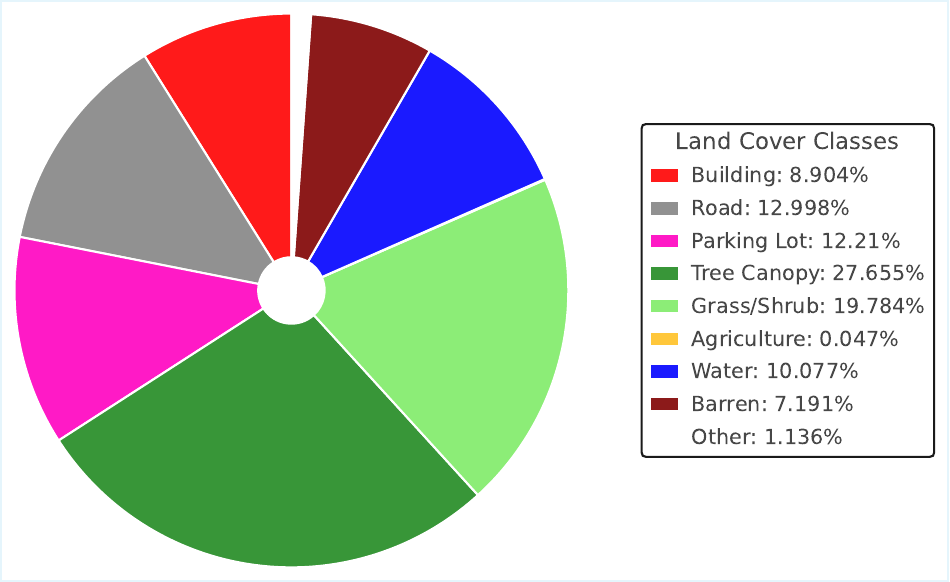}
    \caption{Visualization of land cover class distribution.}
    \label{fig:class_distribution}
\end{figure}

In this study, we introduce the \textit{Baltimore Atlas Dataset}, an ultra-high-resolution remote sensing segmentation dataset constructed from 0.3 m aerial imagery of Baltimore City acquired through the National Agriculture Imagery Program (NAIP). The imagery is annotated into nine urban semantic categories defined by \textit{UrbanWatch} through a two-stage process. In the first stage, we use the \textit{UrbanWatch} model, trained on 1 m resolution imagery, to generate the initial annotations as a coarse baseline. In the second stage, these annotations are refined through manual corrections to improve accuracy. For sampling, we extract 1024 × 1024 pixel patches (300 m × 300 m) from the entire imagery to meet SAM2’s input requirements, while applying uniform sampling to maintain class balance. This process yields 60 samples (62,914,560 pixels in total), with the class distribution shown in Fig.~\ref{fig:class_distribution}. Among them, 55 are used for training and 5 for testing, providing enough scale and diversity for effective model training and reliable performance evaluation.

\begin{figure*}[t]
    \centering
    \includegraphics[width=1\textwidth]{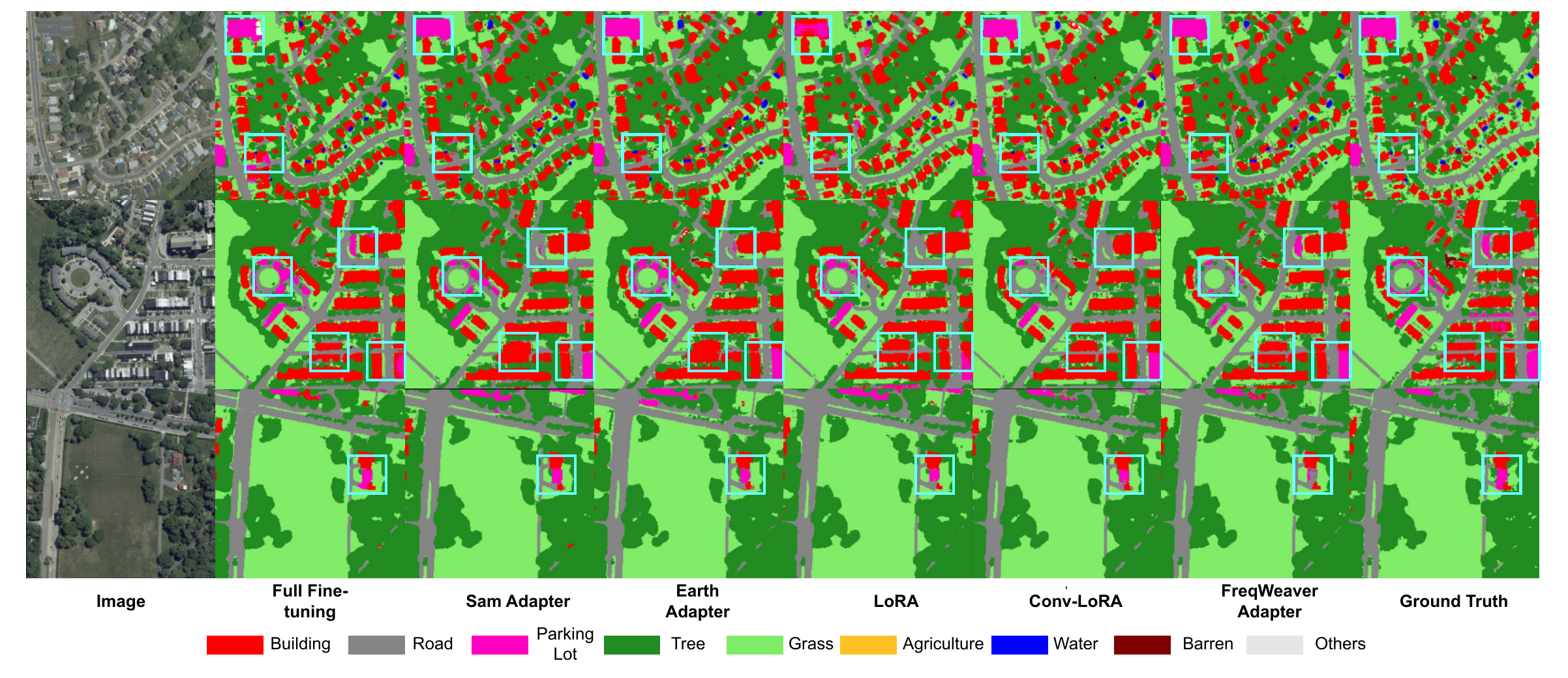} 
    \caption{Visualization of the State-of-the-Art PEFT Method. In the first row, our method shows better capture of textures and boundaries of small objects in dense urban areas, such as residential swimming pools. In the second and third rows, parking lots are recognized at the level of individual vehicles. Although this differs from region-level annotations, it highlights the strength of our local detail modeling and suggests directions for future fine-grained classification.}
    \label{fig:peft_vis}
\end{figure*}

\subsection{Evaluation Metrics}

To comprehensively evaluate the segmentation performance, we adopt two widely used metrics: Intersection over Union (IoU) and Dice coefficient. Both metrics quantify the region-level overlap between predicted segmentation results and ground truth annotations, but with different emphases: IoU focuses on overall overlap consistency, while Dice is more sensitive to small objects and class imbalance. In combination, these metrics provide a comprehensive and complementary assessment of segmentation accuracy.

Given the predicted region \( P \) and the ground truth region \( G \), the IoU and Dice coefficient are defined as follows:
\begin{equation}
\text{IoU} = \frac{|P \cap G|}{|P \cup G|},
\end{equation}
\begin{equation}
\text{Dice} = \frac{2|P \cap G|}{|P| + |G|},
\end{equation}
where \( |P \cap G| \) denotes the number of pixels in the intersection of the prediction and the ground truth, and \( |P \cup G| \) denotes the number of pixels in their union. Higher values of IoU and Dice indicate better segmentation performance.

\subsection{Training and Testing Details}
All experiments are conducted using PyTorch on a single NVIDIA RTX L40S GPU with 48~GB of memory. In our work, we adopt \texttt{SAM2} as the backbone network. Input images are uniformly resized to $1024 \times 1024$ during both training and testing. The model is trained for 200 epochs using the Adam optimizer with a learning rate of $5 \times 10^{-4}$ and a batch size of 2. 

For the segmentation task, we employ the standard Cross-Entropy loss, placing emphasis on accurate pixel-wise classification as the primary optimization objective.

\begin{table}[htbp]
\centering
\caption{Quantitative Comparison with State-of-the-Art Parameter-Efficient Fine-Tuning (PEFT) Methods}
\label{tab:adapter_comparison_peft}
\resizebox{0.5\textwidth}{!}{%
\begin{tabular}{c|cc|c}
\toprule
\textbf{Methods} & \textbf{IoU} & \textbf{Dice} & \textbf{Param} \\
\midrule
Full Fine-tuning              & 0.5239 & 0.6006 & 69.11M \\
Earth Adapter (2025) \cite{hu2025earthadapterbridgegeospatialdomain}          & 0.5149 & 0.5893 & 10.19M \\
LoRA (2021) \cite{hu2021loralowrankadaptationlarge}                  & 0.5202 & 0.5909 & \textbf{0.48M} \\
SAM Adapter (2023) \cite{wu2023medicalsamadapteradapting}    & 0.5321 & 0.6116 & 5.64M \\
Conv-LoRA (2024) \cite{zhong2024convolutionmeetsloraparameter}             & 0.5342 & 0.6051 & 0.50M \\
\textbf{Ours}          & \textbf{0.5520} & \textbf{0.6331} & 4.12M \\
\bottomrule
\end{tabular}
}
\end{table}

\begin{figure*}[t]
    \centering
    \includegraphics[width=1\textwidth]{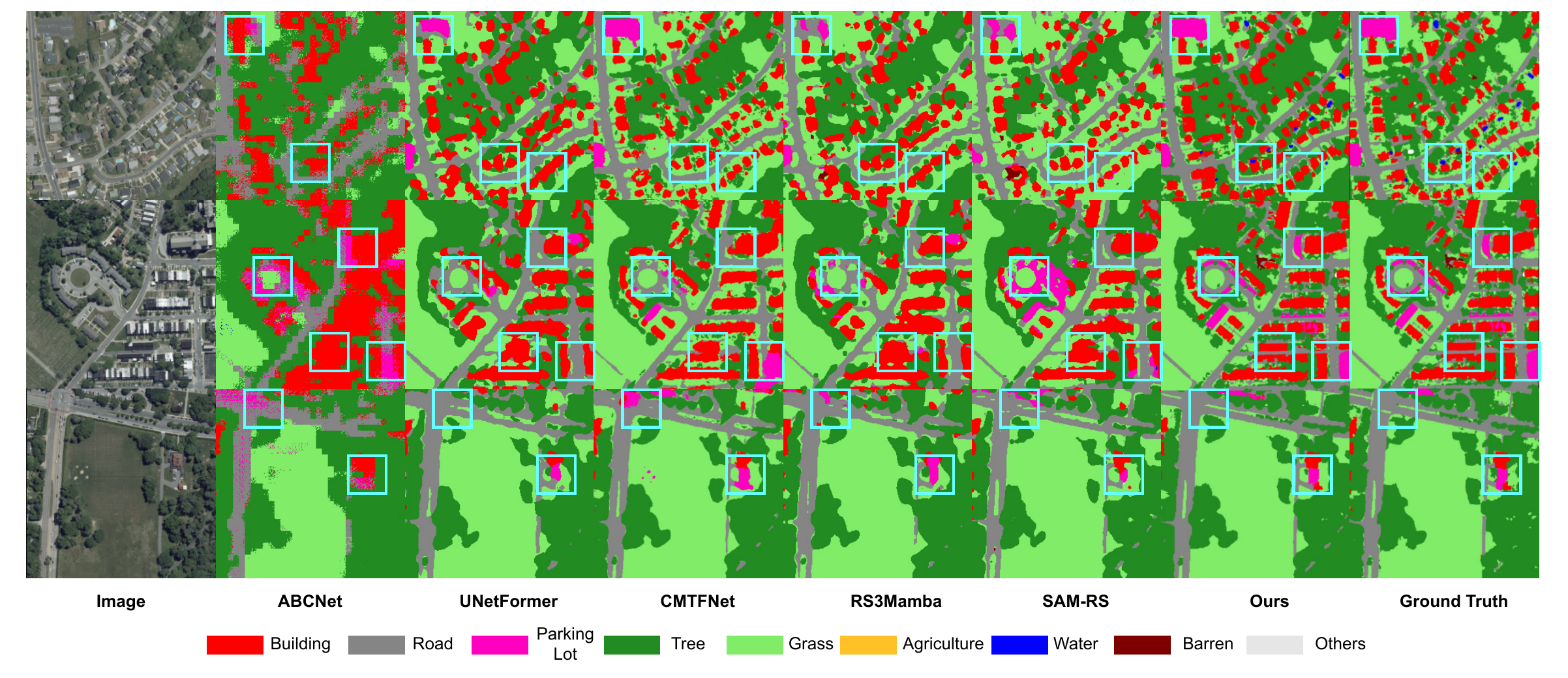} 
    \caption{Visualization of the State-of-the-Art LULC Method. The results show that earlier methods do not work well on ultra-high spatial resolution data: (1) they have trouble outlining building boundaries, for example, nearby structures are often blurred together; (2) they miss small objects, such as residential swimming pools; and (3) they mix up similar classes, for instance confusing roads with parking lots. In contrast, our method gives clear improvements on all these points.}
    \label{fig:lulc_vis}
\end{figure*}

\subsection{Comparison With State-of-the-Art PEFT Methods}

To evaluate the effectiveness of the proposed adapter, we conduct extensive comparative experiments against several state-of-the-art parameter-efficient fine-tuning (PEFT) methods, including the SAM Adapter~\cite{wu2023medicalsamadapteradapting}, LoRA~\cite{hu2021loralowrankadaptationlarge}, Conv-LoRA~\cite{zhong2024convolutionmeetsloraparameter}, and the Earth Adapter~\cite{hu2025earthadapterbridgegeospatialdomain}. All implementations are based on SAM2, ensuring a consistent backbone for fair comparison. These approaches represent diverse strategies for adapting SAM2, with different focuses on domain specificity and parameter efficiency. The comparison results are shown in Table~\ref{tab:adapter_comparison_peft}.

\textbf{Limitations of LoRA-Based Methods.}
Our experimental results show that both LoRA and Conv-LoRA achieve competitive performance with very low parameter overhead. LoRA leverages low-rank decomposition for efficient adaptation and is widely used in lightweight fine-tuning tasks. However, due to its linear structure, it has limited ability to capture the spatial complexity of high-resolution remote sensing scenes, which often contain multi-scale objects, intricate semantics, and fine-grained textures. To mitigate this limitation, Conv-LoRA extends the original LoRA design by incorporating convolutional layers to enhance local feature extraction. This modification improves its capacity to model small objects and boundary details, resulting in measurable performance gains. Nevertheless, despite these improvements, Conv-LoRA still falls short in fully exploiting the potential of high-resolution remote sensing imagery, particularly in capturing multi-scale spatial structures and fine-grained textures, which indicates the need for further refinement.

\textbf{Limitations of Adapter-Based Methods.}
The SAM Adapter is originally developed for medical image segmentation and shows a certain level of adaptability when transferred to remote sensing tasks. However, medical images usually contain relatively simple structures with clear boundaries, while remote sensing data have higher resolution, more complex scene variations, and richer semantic categories. As a result, the SAM Adapter has limited ability to model fine-grained structures and semantic differences, and its performance improvement remains modest. In contrast, the Earth Adapter is specifically designed for remote sensing tasks, but its attempt to decouple frequency components lacks targeted high- and low-frequency modeling mechanisms. This limitation prevents it from fully capturing complex spatial structures and texture features, leading to higher parameter overhead and unsatisfactory overall performance.

\textbf{Effectiveness of FreqWeaver Adapter.}
Our FreqWeaver Adapter introduces a frequency-aware design for remote sensing applications. It explicitly decouples low- and high-frequency components and applies targeted processing to each, which suppresses low-frequency artifacts while preserving high-frequency structural details. This improves segmentation accuracy and allows better adaptation to the SAM2 backbone. Compared with the best-performing Conv-LoRA, FreqWeaver increases IoU by 1.78\% and Dice by 2.8\%, showing clear gains in both overall segmentation accuracy and fine-grained detail modeling. In addition, FreqWeaver achieves this higher performance with fewer parameters than both the Earth Adapter and the SAM Adapter, demonstrating strong domain-specific effectiveness and efficiency for high-resolution remote sensing tasks. We also provide visualizations to further validate its effectiveness, as shown in Fig.~\ref{fig:peft_vis}.

\begin{table}[t]
\centering
\caption{Quantitative Comparison with state-of-the-art Land Use and Land Cover (LULC) methods.}
\label{tab:adapter_comparison_lulc}
\begin{tabular}{c|cc}
\toprule
\textbf{Methods} & \textbf{IoU} & \textbf{Dice} \\
\midrule
ABCNet(2021)\cite{Li_2021} & 0.4111 & 0.5217 \\
UNetFormer(2022)\cite{WANG2022196} & 0.5236 & 0.6201 \\
CMTFNet(2023)\cite{10247595} & 0.5428 & 0.6369 \\
RS$^3$Mamba(2024)\cite{10556777} & 0.5092 & 0.6068 \\
SAM-RS(2024)\cite{10636322} & 0.5434 & 0.6342 \\
\textbf{Ours} & \textbf{0.5778} & \textbf{0.6545} \\
\bottomrule
\end{tabular}
\end{table}

\subsection{Comparison with State-of-the-Art LULC Methods}

To evaluate the effectiveness of our proposed approach, we conduct extensive comparative experiments against several representative state-of-the-art Land Cover and Land Use Classification (LULC) methods. Specifically, we select five recent and competitive baselines from the literature (2021–2024)—ABCNet~\cite{Li_2021}, UNetFormer~\cite{WANG2022196}, CMTFNet~\cite{10247595}, RS$^{3}$Mamba~\cite{10556777}, and SAM-RS~\cite{10636322}—chosen for their strong benchmark performance and representation of diverse architectural paradigms, including CNN-based, Transformer-based, and Mamba-based models. The quantitative comparison results are shown in Table~\ref{tab:adapter_comparison_lulc}, while qualitative visualization examples are presented in Fig.~\ref{fig:lulc_vis} to further illustrate the advantages of our method.

\textbf{Limitations of Existing Methods.}  
Quantitative results (Table~\ref{tab:adapter_comparison_lulc}) demonstrate that the proposed method consistently outperforms all selected baseline methods. In comparison to the strongest baseline, SAM-RS, our model achieves an improvement of 3.44\% in IoU and 2.03\% in Dice score, underscoring its superior performance in LULC tasks. Despite the reasonable segmentation accuracy exhibited by prior methods, visual analysis reveals several limitations:  
(1) Misclassification remains prevalent in complex scenes, particularly those with high intra-class variance or subtle inter-class distinctions (e.g., differentiating various types of vegetation or urban structures);  
(2) Object boundaries are often blurry and imprecise, a critical deficiency for ultra-high-resolution segmentation tasks that demand sharp delineations;  
(3) Models generally perform well on spectrally distinct classes like grass and trees, but falter on categories with complex geometries or semantic overlaps, such as distinguishing between parking lots and buildings, or between roads and barren land.  
Moreover, small object detection, such as identifying pools in residential areas, is often overlooked by existing approaches.

\textbf{Effectiveness of Our Method.}  
We adapt the SAM2 foundation model for the LULC task, leveraging its rich prior knowledge to obtain robust category- and pixel-level representations. Building upon this foundation, our FreqWeaver Adapter enhances the semantic granularity of feature modeling by explicitly capturing and integrating multi-frequency information crucial for detail preservation, thereby enabling significantly improved delineation of object boundaries and detection of small structures. Notably, the enhanced feature representation from the FreqWeaver Adapter allows our model to effectively exploit contextual cues to better distinguish between semantically similar regions such as parking lots and roads. These capabilities demonstrate the strengths of our architecture in ultra-high-resolution LULC tasks. In addition, the integration of a semi-supervised learning strategy further improves the robustness and generalization of our model under limited annotation scenarios.

\begin{table}[ht]
\centering
\caption{Classification performance on ultra-high resolution Downstream Applications.}
\resizebox{0.98\linewidth}{!}{%
\begin{tabular}{lcc}
\toprule
\textbf{Class} & \textbf{User's Accuracy (\%)} & \textbf{Producer's Accuracy (\%)} \\
\midrule
Building       & 93.71 & 81.71 \\
Road           & 85.39 & 89.94 \\
Parking lot    & 85.00 & 85.53 \\
Tree Canopy    & 75.58 & 98.98 \\
Grass/Shrub    & 91.80 & 73.36 \\
Agriculture    & 0.00  & 0.00  \\
Water          & 94.92 & 98.25 \\
Barren         & 80.00 & 40.00 \\
Other          & 60.00 & 54.55 \\
\midrule
\textbf{Overall Accuracy (OA)}  & \multicolumn{2}{c}{\textbf{84.55\%}} \\
\textbf{Kappa Coefficient}      & \multicolumn{2}{c}{\textbf{81.38\%}} \\
\bottomrule
\end{tabular}%
}
\label{tab:map_ana}
\end{table}

\begin{figure}[t]
    \centering
    \includegraphics[width=0.5\textwidth]{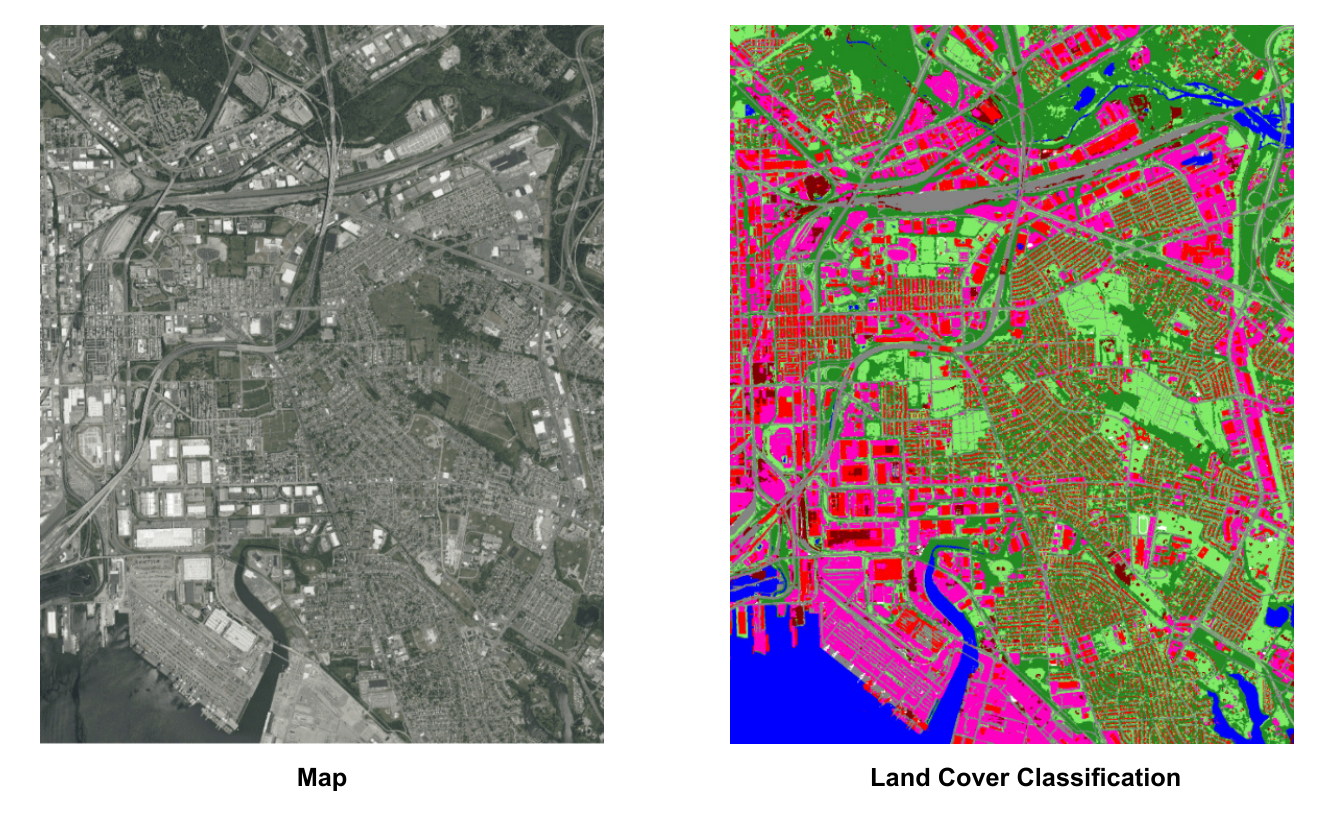}
    \caption{Visualization of Land Cover Classification on the Baltimore City Map. The results show that our method achieves reliable land cover classification across the entire city at ultra-high spatial resolution, clearly revealing the overall urban structure.}
    \label{fig:map_result}
\end{figure}

\subsection{Downstream Applications in Ultra-High Spatial Resolution Map Analysis}

To evaluate the practicality of our approach in real-world scenarios, we apply the proposed method to downstream ultra-high-resolution map analysis tasks, as summarized in Table~\ref{tab:map_ana}. Despite being trained on a relatively small dataset, our method exhibits strong generalization capabilities, achieving an Overall Accuracy (OA) of 84.55\% and a Kappa coefficient of 81.38\%. These metrics reflect not only the correctness of predictions but also the statistical reliability of classification performance across diverse land cover categories.

A detailed breakdown of class-wise performance is shown in Table~\ref{tab:map_ana}. Notably, our model achieves high User’s Accuracy for categories such as \textit{Building} (93.71\%), \textit{Water} (94.92\%), and \textit{Grass/Shrub} (91.80\%), indicating precise identification of spatial regions with distinct spectral or structural characteristics. At the same time, it maintains balanced Producer’s Accuracy, particularly for \textit{Tree Canopy} (98.98\%) and \textit{Road} (89.94\%), demonstrating strong recall and comprehensive spatial coverage.

Nevertheless, certain challenges persist in distinguishing semantically overlapping or visually ambiguous regions, such as \textit{Tree Canopy vs. Grass/Shrub} and \textit{Parking Lot vs. Road}, which are frequently misclassified due to spatial proximity or similar textures.

To further validate the effectiveness of our method, we perform qualitative visualizations on real-world map segments, as illustrated in Fig.~\ref{fig:map_result}. The results show that our approach successfully preserves object boundaries, captures fine-grained structures, and adapts effectively to spatial heterogeneity, confirming its suitability for downstream high-resolution map analysis tasks.

\subsection{Ablation Studies}
To evaluate the contribution of each component within the proposed framework, we conduct extensive ablation studies, as shown in Table~\ref{tab:ablation_Quantitative}. In each experiment, specific modules—FreqWeaver (FW) and the Uncertainty-Aware Teacher–Student Framework (UATS)—are selectively enabled or disabled to analyze their individual impact on segmentation performance.

\textbf{Effectiveness of FreqWeaver.}  
We introduce FreqWeaver (FW) to decouple the frequency components and apply specific modeling within SAM2, enabling a parameter-efficient mechanism to capture complex patterns in remote sensing imagery. Experimental results show that, compared to full fine-tuning, our method achieves a 2.81\% improvement in IoU and a 3.25\% improvement in Dice score, while updating only 5.96\% of the total parameters. These findings highlight the effectiveness of FreqWeaver in fine-tuning large vision models for remote sensing tasks and provide a scalable pathway for adapting foundation models under strict resource constraints.

\textbf{Effectiveness of the Uncertainty-Aware Teacher–Student Model.}  
We introduce the Uncertainty-Aware Teacher–Student model (UATS) to mitigate the impact of unreliable pseudo-labels, which are common in remote sensing imagery due to complex backgrounds and ambiguous boundaries. By incorporating uncertainty estimation into the consistency learning framework, UATS effectively downweights noisy supervision and makes better use of unlabeled data in the semi-supervised setting. As shown in our experiments, adding UATS on top of FreqWeaver leads to a 2.58\% improvement in IoU and a 2.14\% improvement in Dice score. These results demonstrate the importance of uncertainty modeling in enhancing model robustness and generalization across diverse remote sensing scenes.

\begin{table}[t]
\centering
\caption{Quantitative results of the ablation study.}
\label{tab:ablation_Quantitative}
\resizebox{0.6\linewidth}{!}{%
\begin{tabular}{cc|cc}
\toprule
\textbf{FW} & \textbf{UATS} & \textbf{IoU} & \textbf{Dice} \\
\midrule
\ding{55} & \ding{55} & 0.5239 & 0.6006 \\
\ding{51} & \ding{55} & 0.5520 & 0.6331 \\
\ding{51} & \ding{51} & \textbf{0.5778} & \textbf{0.6545} \\
\bottomrule
\end{tabular}
}
\end{table}

\begin{table}[htbp]
\centering
\caption{Quantitative results of the Adaptation ablation study.}
\label{tab:Adaption_Quantitative}
\begin{tabular}{cccc|cc}
\toprule
\textbf{Stage 1} & \textbf{Stage 2} & \textbf{Stage 3} & \textbf{Stage 4} & \textbf{IoU} & \textbf{Dice} \\
\midrule
\ding{51} & \ding{51} & \ding{51} & \ding{51} & 0.5238 & 0.5997 \\
\ding{55} & \ding{51} & \ding{51} & \ding{51} & 0.4955 & 0.5694 \\
\ding{55} & \ding{55} & \ding{51} & \ding{51} & \textbf{0.5520}  & \textbf{0.6331} \\
\ding{55} & \ding{55} & \ding{55} & \ding{51} & 0.3722 & 0.4633 \\
\bottomrule
\end{tabular}
\end{table}

\subsection{Segment Anything 2 Adaptation}

Segment Anything 2 is built upon the Hierarchical Detection Transformer (HieraDet), a multi-scale transformer architecture composed of four stages. In the original Segment Anything framework, adapters such as the SAM Adapter are inserted into every transformer block. However, directly applying this strategy to Segment Anything 2 often results in suboptimal fine-tuning performance due to its deeper and hierarchically structured design. To better accommodate these architectural differences, we conduct a series of empirical studies to investigate more effective adapter integration strategies. The results are shown in Table~\ref{tab:Adaption_Quantitative}.

Specifically, we hypothesize that fine-tuning a foundation model to adapt to downstream tasks such as remote sensing segmentation primarily requires the adjustment of deeper semantic layers. Therefore, we initially insert adapters into all four stages and gradually move toward inserting them only in the deeper stages. Experimental results reveal that adding adapters to early transformer stages not only fails to improve domain adaptation but can even hinder the model's ability to specialize for remote sensing data. Conversely, limiting adapter placement to only the final stage also leads to underfitting, as the model lacks sufficient capacity for structural adaptation. We observe that inserting adapters into Stage 3 and Stage 4 yields the best performance, striking a balance between semantic depth and structural flexibility. While our experiments primarily report results using the proposed FreqWeaver Adapter, we also validate this stage-wise trend using other adapter designs, including the SAM Adapter and Earth Adapter, and find consistent behavior across all cases.

\section{Conclusion}
In this work, we present Baltimore Atlas, a framework that combines the FreqWeaver Adapter and the Uncertainty-Aware Teacher Student Framework for ultra-high spatial resolution land cover classification. The FreqWeaver Adapter separates image information into high- and low-frequency components, modeling each to capture both global and local details, and adapts SAM2 to UHSR data with minimal parameters. The Uncertainty-Aware Teacher–Student Framework uses high-confidence pseudo-labels to learn more reliable features from unlabeled data. Combined, these components support fine-grained UHSR land cover classification and reduce the need for large labeled datasets. Experiments on the Baltimore Atlas dataset show that our FreqWeaver Adapter outperforms leading parameter-efficient tuning methods, and the full framework achieves better results than current state-of-the-art land cover classification approaches.

\section{Discussion}
One limitation of our current study lies in the geographic scope of the dataset, which is primarily extracted from selected regions within Baltimore City. Although class balancing techniques were applied during preprocessing, certain land cover categories—such as agriculture—remain significantly underrepresented. This imbalance may introduce bias into the model’s learning process, particularly in recognizing rare or sparsely distributed classes. In future work, we plan to extend our dataset to comprehensively cover the entire Baltimore City area. By doing so, we aim to achieve a more balanced and representative distribution of land cover classes, which is expected to enhance the generalizability and robustness of the segmentation model.

\begin{acks}
This research is part of the funded project "BaltimoreAtlas: An ultra-high-resolution land cover and land use classification for the Greater Baltimore Metropolitan Area", supported by the School of Emerging Technologies (SET) seed funding, awarded by Towson University. Additional data acquisition, processing, and analysis were supported by the U.S. Geological Survey under Grant/Cooperative Agreement No. G23AP00683 (GY23-GY27).
\end{acks}

\bibliographystyle{ACM-Reference-Format}
\bibliography{paper}

\appendix

\end{document}